\documentclass{ieeeaccess}
\usepackage{cite}
\usepackage{amsmath,amssymb,amsfonts}
\usepackage{algorithmic}
\usepackage{algorithm}
\usepackage{diagbox}
\usepackage{graphicx}
\usepackage{textcomp}
\def\BibTeX{{\rm B\kern-.05em{\sc i\kern-.025em b}\kern-.08em
    T\kern-.1667em\lower.7ex\hbox{E}\kern-.125emX}}
\begin{document}
\history{Date of publication xxxx 00, 0000, date of current version xxxx 00, 0000.}
\doi{10.1109/ACCESS.2023.0322000}

\title{VeML: An End-to-End Machine Learning Lifecycle for Large-scale and High-dimensional Data}
\author{\uppercase{Van-Duc Le}\authorrefmark{1},
\uppercase{Tien-Cuong Bui}\authorrefmark{2}, and Wen-Syan Li\authorrefmark{3}}

\address[1]{Department of Electrical and Computer Engineering, Seoul National University (e-mail: levanduc@snu.ac.kr)}
\address[2]{Department of Electrical and Computer Engineering, Seoul National University (e-mail: cuongbt91@snu.ac.kr)}
\address[3]{Graduate School of Data Science, Seoul National University (email: wensyanli@snu.ac.kr)}

\markboth
{Le \headeretal: VeML: A Version Management System for End-to-End Machine Learning Lifecycle}
{Le \headeretal: VeML: A Version Management System for End-to-End Machine Learning Lifecycle}

\corresp{Corresponding author: Van-Duc Le (e-mail: levanduc@snu.ac.kr).}

\begin{abstract}
An end-to-end machine learning (ML) lifecycle consists of many iterative processes, from data preparation and ML model design to model training and then deploying the trained model for inference. When building an end-to-end lifecycle for an ML problem, many ML pipelines must be designed and executed that produce a huge number of lifecycle versions. Therefore, this paper introduces VeML, a Version management system dedicated to end-to-end ML Lifecycle. Our system tackles several crucial problems that other systems have not solved. First, we address the high cost of building an ML lifecycle, especially for large-scale and high-dimensional dataset. We solve this problem by proposing to transfer the lifecycle of similar datasets managed in our system to the new training data. We design an algorithm based on the core set to compute similarity for large-scale, high-dimensional data efficiently. Another critical issue is the model accuracy degradation by the difference between training data and testing data during the ML lifetime, which leads to lifecycle rebuild. Our system helps to detect this mismatch without getting labeled data from testing data and rebuild the ML lifecycle for a new data version. To demonstrate our contributions, we conduct experiments on real-world, large-scale datasets of driving images and spatiotemporal sensor data and show promising results.
\end{abstract}


\titlepgskip=-21pt

\maketitle

\section{Introduction}
\label{sec:introduction}

\begin{figure}[ht]
  \centering
  \includegraphics[width=\linewidth]{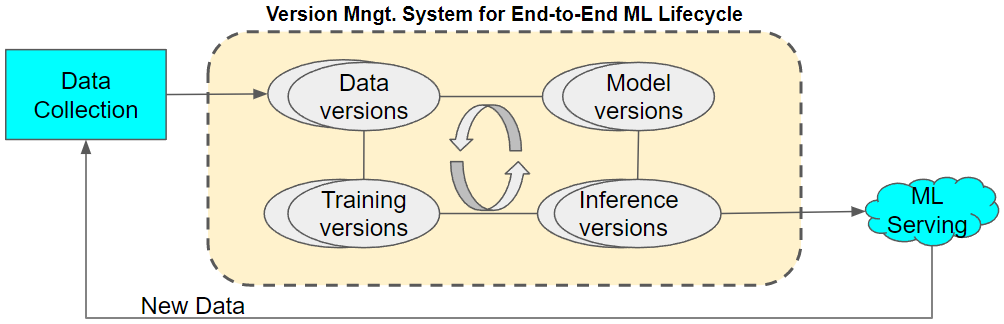}
  \caption{Data flow of our version management system for end-to-end ML lifecycle.}
  \label{fig:version_mngt_system}
\end{figure}

\PARstart{F}{irstly}, we try to answer the question: why do we need a version management system for the end-to-end ML lifecycle? When building an end-to-end ML lifecycle, we need to deal with many possible choices for data preparation, ML algorithms, training hyper-parameters, and deployment configurations. As a results, it costs huge time and computation to build an end-to-end ML lifecycle. Moreover, the ML task continuously evolves throughout its lifetime that produces a a lot of lifecycle versions, from data versions to inference versions. Therefore, we built our \underline{V}ersion management system dedicated to the \underline{e}nd-to-end \underline{ML} lifecycle (VeML) to manage many ML lifecycle versions and leverage the stored versions for efficiently building a new ML lifecycle. Figure \ref{fig:version_mngt_system} shows the data flow of our system from the data collection through our ML version management to model serving and go back with the new data.

In this paper, we raise some crucial research questions for an end-to-end ML lifecycle management system that existing systems do not fully solve. We will show that our proposed VeML system can tackle these challenges in one unified system.

The first challenge for an ML lifecycle management system is how to manage a huge number of versions in an end-to-end ML lifecycle. Our system is built from ground on an internal in-memory storage engine for large-scale storage, integrating an enterprise-strength graph database like Neo4j \cite{neo4j} for graph-based lifecycle versions management, and a unified ML training framework, OpenMMLab, which supports from data preparation to model deployment \cite{openmmlab}. Therefore, our system can manage large-scale datasets and can support end-to-end ML lifecycle versions, from data to inference versions.

The second challenge deals with the problem of how to leverage a large number of historic ML lifecycle versions to efficiently build an ML lifecycle for a new ML application. Especially, this challenge raises two research questions: How to save time and computation in building an ML pipeline for a new training dataset; and How to efficiently retrain for new unseen data during the ML lifecycle. We illustrate the huge cost of building an end-to-end ML lifecycle through the object detection problem, which is an important ML task for many real-world applications.

The training dataset for an object detection problem is often in large-scale. For example, the detection COCO \cite{mscoco} dataset has more than 120K data samples with the data size is 21GB. The BDD100K \cite{bdd} dataset for diverse driving has 100K object detection frames. To build an ML pipeline for a training data (e.g., COCO dataset), an ML engineer will need to try with many data transformation techniques, ML model algorithms, training hyper-parameters, and inference configurations to achieve the final target (e.g., the highest testing accuracy). We experimented with 4 Nvidia Titan GPUs, each with 24GB GPU memory, then the training time for just one ML pipeline is around 12 hours. The ML engineer can use some automated ML algorithms such as NAS-FCOS \cite{nas-fcos} to automatically find an ML pipeline, but the search cost for a training data is very high, 28 GPU-days, which is inefficient in production.

Another case is the requirement to rebuild an ML lifecycle when the ML data continuously evolves when the ML problem runs in real-world. This situation is very common for object detection tasks in real-life applications like self-driving car where the autonomous car must deal with new driving cases throughout its lifetime. Therefore, it raises a crucial research question about building a lifecycle for an ML problem: \textit{How can we leverage our VeML system to \textbf{effectively and efficiently build an end-to-end ML lifecycle} for (1) a new training dataset and (2) new testing data during the ML lifetime?}

\textbf{End-to-end ML lifecycle for a training dataset} A training dataset will start a ML pipeline for a new ML problem. To quickly build a lifecycle for the ML problem, we propose the \textit{lifecycle transferring algorithm}, which uses the dataset similarity to transfer lifecycle versions of similar datasets. Our solution is inspired by transfer learning methodology in which we can transfer the whole ML pipeline to a similar dataset to save training time but still get high performance.

The challenge is to efficiently compute dataset similarity for large-scale, high-dimensional data. ML datasets are often high dimensions (e.g., 1280x720 image data) and consist of large samples (e.g., COCO, BDD datasets have more than 100K examples). Thus, it is very inefficient to compute dataset similarity using all data samples of each dataset. To solve it, we propose representing each dataset as a small core set that can cover its distribution to efficiently compute similarity for each pair of datasets in the VeML system.

\textbf{End-to-end ML lifecycle for new testing data} A new testing data is a collection of unseen data samples when the ML problem runs in the real-world production. As a result, new testing data continuously come during the ML lifetime. A drift testing data is a data version that causes the (deployed) model accuracy significantly drops. The drift testing data version is derived from a different distribution than the training data version. If the testing and training data version are drawn from the same data distribution, no model accuracy degradation occurs; thus, the ML lifecycle remains. On the other hand, retraining is needed, then we need to construct a new ML lifecycle for the new testing data version. 

In this paper, we propose to compare the core set of both testing and training data versions to detect data distribution mismatch without getting labeled test data, which is human cost saving. The next challenge is how to efficiently rebuild an ML lifecycle for a new testing data version in the case of the data distribution difference. We achieve this by allowing ML engineers to choose from various incremental training methods and VeML will automatically rebuild a new ML lifecycle after that.

In summary, we present our contributions for this research as follows:
\begin{itemize}
    \item We build a version management system dedicated to end-to-end ML lifecycle (VeML), from data to inference. Our system implements numerous functionalities to help manage huge ML lifecycle versions.
    \item We propose an algorithm based on the core set to efficient comparing large-scale and high-dimensional data versions. We prove our solution on large-scale driving images and spatiotemporal sensor datasets.
    \item Using dataset similarity computation, our system can transfer lifecycle versions of similar datasets to effectively and efficiently build an ML lifecycle for a new ML problem.
    \item We employ the core set computation to detect data distributions dissimilarity between the testing and training data versions without getting labeled data. Based on the unsupervised data distribution mismatch detection, VeML can support automatically rebuild a ML lifecycle after choosing a model retraining method.
    \item Moreover, to demonstrate that our system is helpful, we show how VeML is using in an on-going self-driving project and how it supports new challenges in ML lifecycle.
\end{itemize}

The rest of this paper is structured as follows. Section 2 presents related research to our work. Section 3 describes our system architecture and functionalities in detail. Section 4 presents how to transfer ML lifecycle versions for a new training dataset. Next, section 5 shows how to detect data distribution mismatch and rebuild a new ML lifecycle. Then, section 6 demonstrates the usefulness of our VeML system. And finally, section 7 wraps up our contributions and discusses future work.

\begin{figure*}
    \centering
    \includegraphics[width=\textwidth]{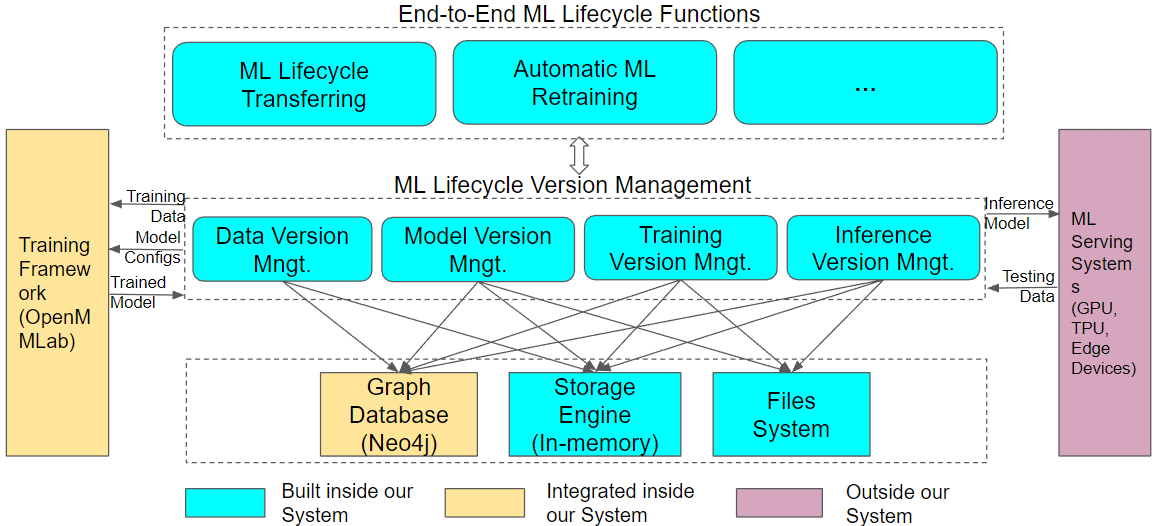}
    \caption{System Architecture and Functionalities.}
    \label{fig:system_architecture}
\end{figure*}

\section{Related Work}
This section discusses related research in ML lifecycle platforms, version control systems, and ML automation which directly connects to our research. We also survey papers tackling data-related challenges, such as dataset similarity, data drift detection, and incremental training with new data.

\textbf{ML Lifecycle Platforms} Many ML lifecycle platforms have been proposed to support ML tasks in production. One of the first such systems is Google Tensorflow Extended (TFX) \cite{tfx}, which has been introduced since 2017. TFX is a TensorFlow-based ML platform, from data preparation to model training and production serving. The versioning information is managed by a metadata tool and can be saved to a database like SQLite or MySQL. MLFlow \cite{mlflow} was presented by DataBricks, the company behind the large-scale data analysis Apache Spark, in 2018. MLFlow is an open-source platform that supports packaging and tracking ML experiments runs and reproducing. It manages ML experiment versions in artifact concepts, such as data files, models, and training codes. Data platform for ML (MLdp) \cite{mldp} was introduced as an internal platform by Apple in 2019. It has an integrated data system, supports data versioning, and integrates with in-house solutions for model training and deployment.

In general, these ML lifecycle platforms do not have end-to-end ML lifecycle version management, from data to inference. In the case of TFX, it supports end-to-end ML lifecycle but does not help build a new ML lifecycle employing managed lifecycle versions as our system.

Recently, \textbf{MLOps for end-to-end ML lifecycle} are emerging. They are provided by many big companies such as Google Cloud \cite{mlops-google}, Amazon Sagemaker \cite{mlops-amazon}, and Microsoft Azure \cite{mlops-microsoft}. These systems support data scientists building end-to-end ML problems, from data to deployment, but still do not leverage many lifecycle versions to quickly construct a lifecycle for an ML problem.

\textbf{Version Management for ML} With the increasing importance of ML versioning management, many solutions have been introduced for ML version control, especially for data versions. Typically, datasets for ML tasks are stored in file systems, causing managing many versions of them difficult and inefficient.

Paper \cite{orpheusdb} proposed to build a data version management system over a relational database. Their solution was to separate the data from the version information in two tables. The data table stores the records are appearing in any data version, while the version table captures the versioning information of which version contains which records. They presented the partitioning optimization problem, given a version-record bipartite graph, minimizing the checkout and storage cost, which is an NP-hard problem.

Our data version management also bases on this idea by separating the data and version storage. We save data samples into in-memory storage but manage the version information in a graph database. Our solution may not optimize the storage cost, but it helps us to load any data versions constantly, which is critical for reproducing any ML training processes during an ML lifecycle.

Moreover, many systems and tools have been proposed to manage data and model versions for the ML lifecycle. For instance, Data Version Control (DVC) \cite{dvc} is a popular open-source tool. DVC lets us capture versions of data and models in Git commits while storing them on-premises or in the cloud. However, no systems supports us in managing end-to-end ML lifecycle versions and leveraging managed versions to build a new ML lifecycle.

\textbf{ML Automation} There are a number of systems that serves automatic searching for the best ML model such as AutoML systems for ML \cite{autosklearn, tpot, kgpip} or NAS systems \cite{nas} for deep learning (DL) problems. These systems search for ML/DL pipelines from a set of predefined ML/DL operators and then execute experiments with many training hyper-parameter combinations. They also leverage similar datasets as a meta-learning approach for more efficient ML pipeline exploration \cite{autosklearn, kgpip}.

The most dissimilarity of these systems to ours is that they search for an ML pipeline for each new dataset, which is time-consuming and high-cost. On the other hand, our system leverages many ML lifecycle versions to effectively and efficiently build new lifecycle for training data and testing data versions.

\textbf{Dataset Similarity} To compute dataset similarity, meta-features based computation is one of the most popular solutions \cite{autosklearn}. However, meta-features are often unavailable for high-dimensional data such as image or spatiotemporal data. Using dataset embedding \cite{kglac-embedding} for dataset similarity computation is also a common method, but it is inefficient when computing with a large number of data samples.

Another recent proposal is computing geometric dataset distances based on optimal transport \cite{geometric}. This method worked for classification datasets but still suffered the high-cost problem when dealing with large-scale datasets. Our similarity computation is based on the core set, a small subset of a dataset, and thus, possible to work with large-scale and high-dimensional datasets.

\textbf{Data Drift Detection} Detecting drift in the continuous data has been tackled in some papers \cite{matchmaker, odin}. Matchmaker \cite{matchmaker} uses a decision tree to detect both data drift and concept drift, but it only works well for tabular data. ODIN \cite{odin} detects drift in video image data, but it still uses all data samples that may not be efficient for massive datasets. Our solution is based on the small core set that can work for unlabeled continuing large-scale data versions.

\textbf{Incremental Learning} continuously retrain an ML model when a new training data comes. Some popular model retraining methods are full training which retrains all available datasets, and transfer learning which only retrains the new dataset from a pre-trained model. These approaches require labeling all available data samples, which is costly. Other incremental learning algorithms, that reduce labeling cost, are active learning \cite{coreset, active-vaal}, which tries to label a small number of the most significant training data, and domain adaptation \cite{adapt1, adapt2, adapt3}, which learns from a source domain but can generalize to a different target domain without labeled data.

\section{System Architecture and Functionalities}

\subsection{System Architecture}

Our system architecture has three main blocks and other functional modules. The first is an in-memory storage engine built in our laboratory to manage large-scale data versions, training logs, and metadata information. The second is an integrated graph database such as Neo4j \cite{neo4j} for graph-based ML lifecycle version management and analysis. And the third component is an ML training framework which is built over the open-source OpenMMLab \cite{openmmlab}.

OpenMMLab is a unified architecture for many ML problems, integrating with common ML frameworks (like PyTorch \cite{pytorch}), easy to re-use and extend functions by a modular design. We leverage the OpenMMLab framework to perform ML training with training data from a data version, model algorithm configurations from a model version, and return trained model checkpoints for a training version. It also supports model deployment to an inference model running in ML serving systems. Figure \ref{fig:system_architecture} shows our system architecture with three main components and many functional modules. We use file systems to save binary objects like trained and deployed models.

\subsection{System Functionalities}

Firstly, we define how we manage the version of every component in the end-to-end ML lifecycle. A \textbf{data version} is a collection of data samples and its data preparation (e.g., normalization, missing values imputation). A \textit{training data version} is a data version that is used as the training data for the ML task. A \textit{testing data version} is a data version that contains the unseen new data collected from the real-world environment when an ML problem runs in production. The unseen test data will be annotated and routed back as training data when rebuilding the ML lifecycle. 

A \textbf{model version} includes a specific ML algorithm (e.g., features transformation, model architecture) to learn from the training data. Different model versions can share some common model structure such as the same model backbone in many object detection algorithms. A \textbf{training version} maintains a set of training hyper-parameters used to optimize the ML model, the training logs, and the trained model. An \textbf{inference version} consists of deployment configurations (e.g., quantization algorithm, inference device) and the deployed model.

The core functionality of our system is the ML lifecycle version management that contains some modules, as shown in figure \ref{fig:system_architecture}. The \textit{data version management} component uses our built in-memory storage engine that can support multiple data types in a unified system, like tabular, image, and graph data. It can filter, update, add, and merge any data versions. It also supports data versions visualization and statistic functions. The \textit{model version module} governs various ML model algorithms as metadata such as model backbone (e.g. ResNet50 \cite{resnet}), ML architecture (e.g. FasterRCNN \cite{fasterrcnn}), and so on. Thus, it provides a model versions comparison function by comparing the metadata of different ML models.

The \textit{training version management} module maintains training hyper-parameters, training logs, and the trained model of each training experiment. It provides training versions visualization and training error analysis functions. The \textit{inference version component} manages deployment configurations and the deployed model of an inference version. It helps to analyze prediction errors by visualizing inference versions on real-world testing data.

Each version management module supplies an application programming interface (API) that accepts a version value and returns the data and metadata information maintained by that component. Therefore, we can build end-to-end ML lifecycle functions over our version management using their APIs (see figure \ref{fig:system_architecture}). This research introduces our implementation for two functions: ML lifecycle transferring and automatic ML lifecycle rebuilding.

Firstly, we implement the \textit{ML lifecycle transferring} function by reusing each lifecycle version, from model to inference, for new training data. Thanks to the APIs of each version management module, it is easy to get each version's data and information and transfer them for the new lifecycle.

Secondly, the \textit{automatic ML lifecycle rebuilding} function is performed by implementing incremental learning methods on the previous lifecycle version. For example, in the full training method, we merge the new testing data version with the previous training data version to be full training data (thanks to our data version management). Then we can reuse the previous model and training versions to train on new training data for a new ML lifecycle.

\begin{figure}
  \centering
  \includegraphics[width=1.0\linewidth]{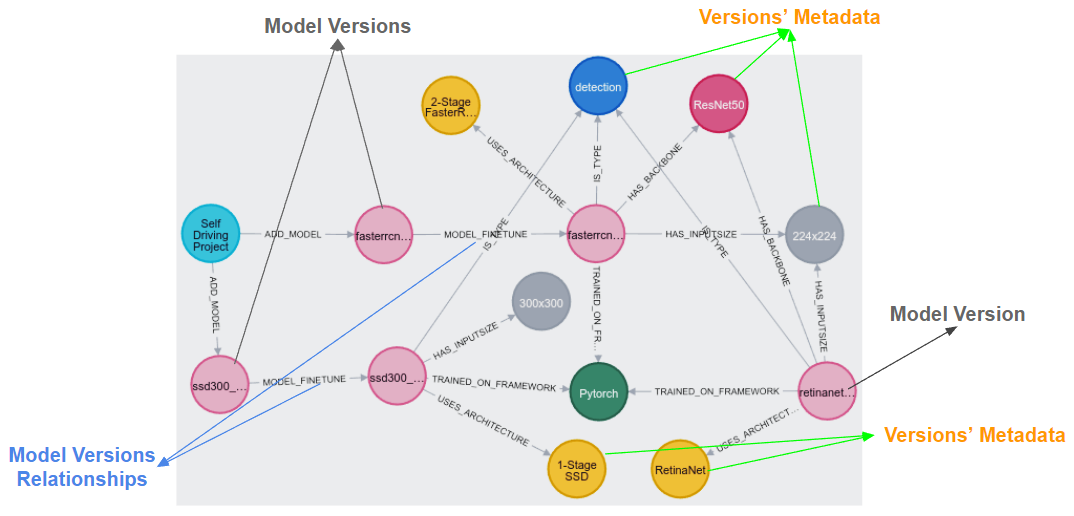}
  \caption{Graph-based management for model versions and model metadata in our system.}
  \label{fig:system_function}
\end{figure}

\subsection{System Implementation} 

Our critical objective is to manage all the data and training logs for numerous end-to-end ML lifecycle versions. We store the data samples, training logs, and other management information in our in-memory storage engine as inserted-only tables without deletion or modification operations. This implementation makes it easier to manage and faster to select. Particularly, we store data records of each data version in one consecutive range of storage that helps to retrieve any data version constantly, which is a benefit for reproducing a training experiment at any time in the ML lifecycle.

Data samples and annotations are stored separately in different tables linked by data samples identifications (data IDs) which are indexed. Thus, it is flexible to manage various types of annotations, such as classes, bounding boxes, segmentation, skeletons, or adding a new one.

Versioning information of a data version is organized in the graph-based schema, with each version being a node in the graph. The set of data IDs for a data version is directly stored in each node, which helps us easily extend or merge any data versions by adding or joining some sets of data IDs.

Moreover, ML configurations like model algorithms and training hyper-parameters are represented as metadata nodes in the graph. All ML versions, like model, training, and inference versions, are also managed in the graph. An ML version with a relationship with each other, such as a model version that is fine-tuned from other model versions, can be denoted as links in our graph management.

Figure \ref{fig:system_function} illustrates how we organize model versions, model metadata, and their relationships in graph-based management. Using graph representation, we can easily inspect an ML lifecycle through any ML version (data to inference) and at any time.

\section{Transferring ML Lifecycle By Dataset Similarity}

In this section, we present how to transfer ML lifecycle versions of a similar dataset in VeML to a new training dataset of a new ML problem. At first, we introduce an efficient dataset similarity solution for large-scale, high-dimensional data based on the core set computation algorithm. Then, we show experimental results on 2 large-scale datasets and discuss the results.

\subsection{Dataset Similarity Challenge}

In this part, we discuss the challenge of datasets similarity computation in recent research. One approach is the meta-features-based method in the AutoML paper \cite{autosklearn}. It computes meta characteristics for each dataset (i.e., statistical features), then ranks all datasets by L1 distance in the meta-feature space, and finally chooses k nearest neighbors. This solution only works for small, tabular datasets but meta-features are not meaningful for high-dimensional data (like image, graph).

The second method is dataset similarity with labels which computes dataset distance by optimal transporting between dataset features and labels \cite{geometric}. The advantage is it can compute different labels datasets (such as between MNIST \cite{mnist} and CIFAR \cite{cifar10}); and leverages both data features and labels. But this method computes using the whole data points so it is inefficient for large-scale datasets like object detection datasets.

The third common solution for dataset similarity computation is to compute distance between data points’ embedding by some distance metrics like L2 distance (Euclidean distance), L1 distance, Gromov–Wasserstein distance \cite{gw-distance} (for different data distribution space). This method works for high-dimensional data and does not depend on data labels. Nevertheless, it still has the problem of highly cost computation with large-scale datasets.

Therefore, we propose an efficient dataset similarity algorithm for large-scale, high-dimension datasets which can work for real-world datasets. Our solution is to compute the core set of each dataset which is a small subset of points that can cover the distribution of the whole dataset \cite{coreset}. Then, we compute the similarity on each pair of datasets in our VeML system as the average distance between their core sets which is efficient in memory and computation.

\subsection{Core Set Computation Algorithm}

For a dataset $A$ with $n$ data samples, we aim to find a small subset $s$ belonging to $A$ with the number of data points in $s$ is $k$ less than $n$, which can represent the distribution of the whole dataset. The subset $s$ satisfied the above condition is called the core set of dataset $A$. In our case, we do not want the core set selection to depend on data labels so we can apply it to any ML problems. Therefore, we follow a similar approach as in the paper \cite{coreset} that chooses a core set of a large dataset based on the embedding of each data sample learned by a Convolutional Neural Network (CNN) model.

Following paper \cite{coreset}, choosing the core set of a dataset $A$ is equivalent to the \textbf{k-center problem}: $min_{A} max_{i} min_{j} \Delta(e_{i},e_{j})$ with $\Delta(e_{i},e_{j})$ = L2-norm distance or Euclidean distance, $e_{i}$ = feature embedding learned by the CNN network for a data point $x_{i}$. 

\begin{figure}[ht]
  \centering
  \includegraphics[width=\linewidth]{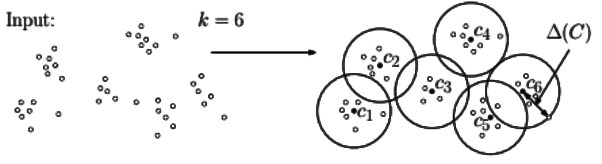}
  \caption{The k-center problem and solution in the Euclidean space, from \cite{coreset-greedy}.}
  \label{fig:core_set_algorithm}
\end{figure}

From the lecture \cite{coreset-greedy}, the k-center problem is stated as follows. Given a set $P$ of $n$ points in a metric space and a number $k$<=$n$, find a set $C$ of $k$ center points to \textit{minimize the maximum distance} of any point of $P$ to its nearest center in $C$. Figure \ref{fig:core_set_algorithm} illustrates the k-center problem and a solution in the Euclidean space. $C$ is the \textit{core set} of $P$, and $\Delta(C)$ is the \textit{covering distance} which is the maximum distance for all points in the data set to its closest center. 

The set of balls established by considering each data point in the core set as the center and the covering distance as the radius is the minimal set of balls that completely covers the distribution of a dataset. They are denoted as \textit{covering balls} of a dataset and are demonstrated in figure \ref{fig:core_set_algorithm} with six covering balls (corresponding to a 6-center core set). Every data point inside the covering balls of a dataset is considered to lie in its data distribution.

\textbf{Greedy algorithm} The k-center problem is NP-hard; therefore, we use a greedy algorithm to approximately compute the core set \cite{coreset-greedy}. The k-center greedy algorithm to find $k$ center points of a dataset $P$ is constructed as follows \cite{coreset-greedy}. The algorithm starts by randomly selecting a point in $P$ as the initial center $g_{1}$. The next center is selected greedily by choosing the point $u$, which is the farthest distance of any point of $P$ from its closet center. This choosing process is repeated until we have $k$ centers.

\begin{figure}[ht]
  \centering
  \includegraphics[width=\linewidth]{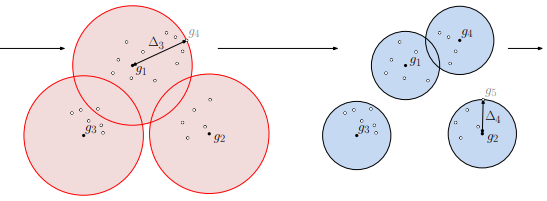}
  \caption{Greedy algorithm to k-center, computes from 3 to 4 centers, from \cite{coreset-greedy}.}
  \label{fig:core_set_greedy}
\end{figure}

$G$ = ${g_{1}, ..., g_{k}}$ is a set of $k$ centers, $\Delta(G)$ = maximum distance of any point of a set of points to its nearest center. It is proved that $\Delta(G)$ <= $2*OPT$ = $2*\Delta(C)$ with $\Delta(C)$ is the optimal solution \cite{coreset-greedy}. Consequently, $G$ is the approximate core set of a dataset, and $\Delta(G)$ is the approximate covering distance. Figure \ref{fig:core_set_greedy} illuminates the greedy algorithm to the k-center problem, computing from 3 to 4 centers.

\subsection{Dataset Similarity Computation}

We propose a dataset similarity computation algorithm for large-scale, high-dimensional data based on the core set. Currently, the number of data samples in an ML dataset is becoming larger and larger. Thus, it is inefficient and expensive to compute dataset similarity using all data points of each data version in our system. Our solution is to select the core set of each data version which is a small subset of data samples that can represent the distribution of the whole data version. We compute the similarity for a pair of datasets as the average distance between their core sets which is efficient in time and computation. Since we cannot compute the exact core set, we use the greedy core set \cite{coreset-greedy} computation as a good approximation.

The dataset similarity computation algorithm procedure is following. Denote $D$ = \{$d_i$\}, $i$ = $1,..,N$ is a dataset, $f$ is the CNN model that will be used to learn the embedding of each data samples in $D$ (e.g., a ResNet50 model \cite{resnet}), $k$ is the number of centers as the approximated core set (e.g., $k$=$10$). The dataset similarity computation between 2 datasets $D_1$ and $D_2$ is illustrated in algorithm \ref{alg:similarity}.

\begin{algorithm}
\begin{algorithmic}
\caption{Dataset similarity computation}\label{alg:similarity}
\REQUIRE $D_1$=\{$d_{1i}$\}, $D_2$=\{$d_{2i}$\}, $f$, $k$
\STATE // Compute the embedding for each dataset
\STATE \{$e_{1i}$\} = $f$(\{$d_{1i}$\}), $i$=$1,..,N_1$
\STATE \{$e_{2i}$\} = $f$(\{$d_{2i}$\}), $i$=$1,..,N_2$
\STATE // Compute k-center by greedy algorithm \cite{coreset-greedy}
\STATE $G_1$ = k-center(\{$e_{1i}$\}, k)
\STATE $G_2$ = k-center(\{$e_{2i}$\}, k)
\STATE // Compute pairwise Euclidean distance
\STATE $d$ = pairwise-distances($G_1$, $G_2$)

\RETURN $d$
\end{algorithmic}
\end{algorithm}

To prove our method, we show the experiment's results by using the greedy approximated core set $G$ to compute the distance between some image datasets. We will test first with small image classification datasets, such as MNIST \cite{mnist}, CIFAR10 \cite{cifar10}, and Fashion MNIST \cite{mnist-fashion} (the numbers of data samples are 60K, 50K and 60K, respectively). We then also test $G$ to use for large-scale, high-dimensional object detection datasets, including MS COCO \cite{mscoco}, BDD \cite{bdd}, KITTI \cite{kitti}, PASCAL-VOC \cite{PascalVoc}, and Cityscapes \cite{cityscapes}.

Table \ref{tab:classification} presents the pairwise distance between classification datasets using two methods. Numbers in the upper corner (italic fonts) are computed using the entire data samples of each dataset. Numbers in the lower corner (normal fonts) are calculated using each dataset's approximated core set $G$. We compute $G$ for every dataset using a 10-center greedy approximation (e.g., k=10).

\textbf{Discussion} From table \ref{tab:classification}, both methods agree that the closest pair of datasets (in bold font) is MNIST and Fashion MNIST. This result is semantic intuition since MNIST and Fashion MNIST are both gray-scale, 28x28 image datasets while CIFAR10 is 32x32 color dataset. Thus, MNIST and Fashion MNIST are more similar than MNIST and CIFAR10. Regarding the memory and computation cost, for dataset similarity computation using all data samples, the memory and computation cost for 2 datasets (e.g., MNIST and CIFAR10) would be 60K x 50K x embedding size (e.g., 1024 bytes). While using k-center core set similarity computation, the memory and computation cost for any pair of datasets would be k x k x embedding size, which is much smaller as k << 50K. Moreover, when we increase the number of centers $k$, the core set based dataset distance is closer to the full data samples distance but the similarity between each dataset is the same so $k$=$10$ can be a good option.

\begin{table}
  \caption{Pairwise dataset distance between classification datasets. \textit{Upper corner numbers are computed using all data points of each dataset. Lower corner numbers are calculated using the core set of each dataset.}}
  \label{tab:classification}
  \begin{tabular}{|p{2.5cm}|p{1.cm}|p{1.cm}|p{2.cm}|}
    \hline
    Dataset & MNIST & CIFAR10 & Fashion MNIST\\
    \hline
    MNIST & & \textit{24.8} & \textit{\textbf{21.0}}\\
    CIFAR10 & 32.2 & & \textit{25.7}\\
    Fashion MNIST & \textbf{28.7} & 34.5 &\\
  \hline
\end{tabular}
\end{table}

We continue our dataset similarity computation experiments by using algorithm \ref{alg:similarity} to compute the distance between large-scale, high-dimensional object detection datasets. Similarly, we use 10-center approximation core set for each dataset. Table \ref{tab:detection} shows the results achieved in our experiments. Distance values in the upper corner (italic fonts) are computed using all data points of each dataset, and values in the lower edge (normal fonts) are obtained using the approximated core set $G$. 

\textbf{Discussion} Table \ref{tab:detection} shows that some datasets like BDD, KITTI, and Cityscapes are closer in the pairwise distance than others. Otherwise, COCO and Pascal VOC datasets are farther to each other and farther than three other datasets. These results are also comparable with computed values using all data samples, that proves our solution. In semantic intuition, these experimental results are reasonable since COCO and Pascal VOC are general object detection datasets, while BDD, KITTI, and Cityscapes are both collected from driving videos. Regarding the memory and computation cost, the k-center core set similarity computation is much more efficient than full data samples computation since $k$ << 100K (the usual data points of an object detection dataset).

Consequently, we can use the k-center approximated core set $G$ to efficiently compute the similarity of large-scale datasets. Our VeML system leverages algorithm \ref{alg:similarity} to compute similarity of each pair of data versions in our system.

\begin{table}[ht]
  \caption{Pairwise distance between object detection datasets. \textit{Upper corner distances are computed using all data samples. Lower edge numbers are calculated using only the core set of each dataset.}}
  \label{tab:detection}
  \begin{tabular}{|p{1.5cm}|p{1.cm}|p{1.cm}|p{1.cm}|p{1.cm}|p{0.5cm}|}
    \hline
    Dataset & COCO & BDD & Cityscapes & KITTI & VOC\\
    \hline
    COCO &  & \textit{15.12} & \textit{13.81} & \textit{14.84} & \textit{15.96}\\
    BDD & 22.45 & & \textit{9.62} & \textit{10.72} & \textit{15.28}\\
    Cityscapes & 21.49 & 12.56 & & \textit{8.23} & \textit{14.24}\\
    KITTI & 22.09 & 13.94 & 10.32 & & \textit{15.17}\\
    VOC & 25.59 & 22.88 & 21.84 & 22.37 & \\
    Our dataset & 21.65 & \textbf{13.14} & \textbf{10.59} & \textbf{12.38} & 21.87\\
  \hline
\end{tabular}
\end{table}

\subsection{ML Lifecycle Version Transferring}

This section presents how we apply the dataset similarity computation to transfer ML lifecycle version to efficiently build end-to-end ML lifecycle for a new ML problem. We start with the a description for the experimental datasets and then show the transferring algorithm and experimental results. We finish with a detailed discussion on the pros and cons of our approach.

\subsubsection{Experimental Datasets}

\begin{figure}[ht]
  \centering
  \includegraphics[width=\linewidth]{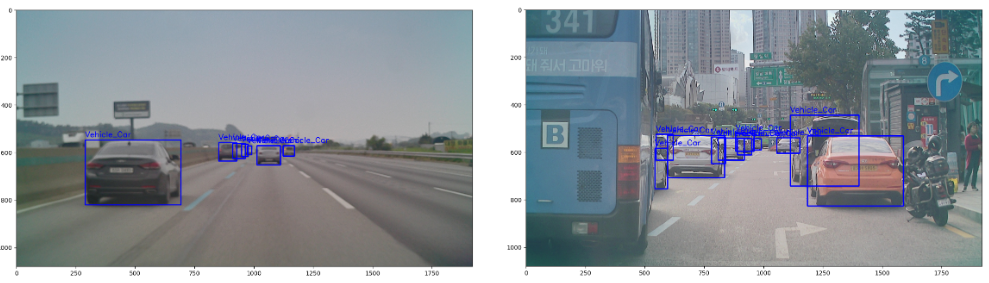}
  \caption{Two data samples with annotations from our real-world image datasets on various driving situations. On the left, driving on a highway. On the right, driving on a city street.}
  \label{fig:dataset_images}
\end{figure}

We examine two types of large-scale, high-dimensional data. The first one is the real-world image dataset of dash cam videos on driving cars in Korea (belongs to a self-driving project) as in figure \ref{fig:dataset_images}. The driving videos were collected in different on-road situations, such as locations, weather, and time of day. The ML problem we experiment with in this dataset is vehicle detection, a critical mission for an autonomous car.

We consider three data versions constructed from driving videos at various conditions to prove our contributions. Table \ref{tab:exp_data_versions} presents three image data versions, their statistics and the collection environment information. Data version D0821 was constructed from 36 driving videos in August 21, 2019, 13h to 17h, on a highway street. Data version D1018 consists of 673 images from 11 videos collected on city streets in Seoul city, Korea, during the afternoon of October 18, 2019. Data version D0114 includes 670 images of 10 driving videos on suburban roadways around Seoul city, from 14h to 16h in January 14, 2020.

\begin{table}
  \caption{Image data versions information}
  \label{tab:exp_data_versions}
  \begin{tabular}{|p{1.cm}|p{1.5cm}|p{1cm}|p{1cm}|p{2.0cm}|}
    \hline
    Data version & Day \& Time & Location & Weather & Statistics\\
    \hline
     D0821 & 08/21/2019 13h-17h & highway & foggy & \#videos:  36 \newline \#images: 1597\\
     D1018 & 10/18/2019 14h-16h & city streets & clear & \#videos: 11  \newline \#images: 673\\
     D0114 & 01/14/2020 14h-16h & suburb streets & overcast & \#videos:  10 \newline \#images: 670\\
  \hline
\end{tabular}
\end{table}

The second experimental data type is the spatiotemporal sensor data, which recently has been getting more attention in ML research. We consider real-world spatiotemporal datasets of traffic speed data (e.g., speed sensors data in LA and Bay Area, USA \cite{dcrnn}) and air pollution data (e.g., PM2.5 and PM10 air pollution data in Seoul, Korea \cite{convlstm}). Our objective is to investigate whether our ML lifecycle transferring algorithm works for spatiotemporal datasets. Specialty, we evaluate the following datasets in our experimentations. 
\begin{itemize}
    \item Dataset \#1: speed sensor data of 207 sensors in LA, USA, 4 months of data.
    \item Dataset \#2: speed sensor data of 325 sensors in the Bay Area, USA, 6 months of data.
    \item Dataset \#3: air pollution data from 25 PM2.5 monitoring stations in Seoul, Korea, 3 years of data.
    \item Dataset \#4: air pollution data of 37 PM10 monitoring stations in Seoul, Korea, 3 years of data.
\end{itemize}
Table \ref{tab:exp_spatiotemporal_data} presents the information of speed sensors data in LA and Bay Area, USA, and air pollution data in Seoul, Korea. The ML problem for all mentioned datasets is a spatiotemporal prediction some time ahead of traffic speed or air pollution.

\begin{table}
  \caption{Spatiotemporal datasets information}
  \label{tab:exp_spatiotemporal_data}
  \begin{tabular}{|p{1.5cm}|p{1.5cm}|p{1cm}|p{1.5cm}|p{1.cm}|}
    \hline
    Dataset & Type & \#Sensors & \#Data points & \#Features\\
    \hline
     Dataset \#1 & Traffic speed & 207 &  6,519,002 & 3\\
     Dataset \#2 & Traffic speed & 325 &  16,937,179 & 3\\
     Dataset \#3 & Air pollution & 25 &  26,280 & 1\\
     Dataset \#4 & Air pollution & 37 &  26,280 & 1\\
  \hline
\end{tabular}
\end{table}

\subsubsection{ML Lifecycle Versions Transferring Algorithm}

To support ML lifecycle transferring with a new training data, we compute the distance between the new one with existing datasets managed in our system via the core set. Thus, we can select top-k of the most similar datasets and transfer their ML lifecycle versions to build a new ML lifecycle. Algorithm \ref{alg:transferring} shows the steps to find the similar datasets in our VeML system for a new training dataset $D$. Denote \{$G_1, G_2,... G_t$\} is a list of k-center core sets for existing datasets in VeML, $f$ is the embedding model.

\begin{algorithm}
\begin{algorithmic}
\caption{ML lifecycle versions transferring algorithm}\label{alg:transferring}
\REQUIRE $D$, \{$G_1, G_2,... G_t$\}, $f$
\STATE // Compute k-center core set for $D$
\STATE $P$ = k-center($f$($D$))
\STATE // Compute pairwise distance between $P$ and \{$G_1, G_2,... G_t$\}
\STATE $d_1$ = pairwise-distances($P$, $G_1$)
\STATE $d_2$ = pairwise-distances($P$, $G_2$)
\STATE ...
\STATE $d_t$ = pairwise-distances($P$, $G_t$)
\STATE // Choose top-k* of the most similar datasets: $d_1$, $d_2$,... $d_{k*}$
\STATE // Execute ML lifecycle versions transferring from top-k* datasets

\RETURN $d_1$, $d_2$,... $d_{k*}$
\end{algorithmic}
\end{algorithm}

\begin{figure}[ht]
  \centering
  \includegraphics[width=\linewidth]{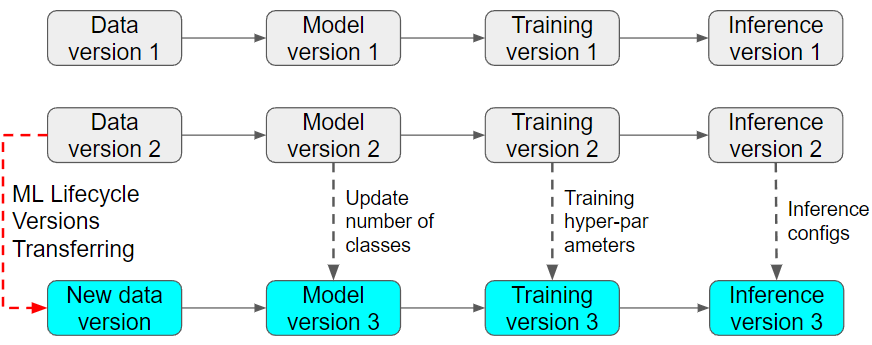}
  \caption{ML lifecycle versions transferring process.}
  \label{fig:lifecycle_transferring}
\end{figure}

Thanks to the modular and configuration-based design of the training framework OpenMMLab \cite{openmmlab}, we can \textit{execute the ML lifecycle versions transferring} as follows. We reuse the configurations of the ML lifecycle versions of the similar dataset, from data preparation to inference configuration, and update specific information following the new training data, such as update the number of classes (e.g., COCO dataset has 80 classes, BDD dataset has 10 classes). Figure \ref{fig:lifecycle_transferring} illustrates the ML lifecycle versions transferring process. Moreover, like transfer learning, we can use the pretrained model managed in VeML to accelerate the training of new ML problem. As a result, since an ML engineer does not have to try with many different ML lifecycle configurations, VeML can \textit{quickly construct a new ML lifecycle for the new ML problem} that shows its \textbf{efficiency}. Then, we will prove in the experimental section the \textbf{effectiveness} of this solution in producing high model accuracy for large-scale, high-dimensional datasets.

\textbf{Experimental setup} We compile and install our system on a server with an AMD EPYC 7502 CPU @ 2.5GHz with 32 cores and 505GB of RAM, running Ubuntu 18.04 LTS. We use Neo4j community version 4.2.18 and OpenMMLab version 1.6.0. We implement all experiments on Pytorch 1.12.1 and run on 4 TITAN RTX GPUs, each has 24GB of RAM.

\subsubsection{Object Detection Transferring Experiments} 

In this section, we prove our ML lifecycle transferring solution for object detection problem with common object detection datasets as well as our real-world driving image dataset. We will validate how ML lifecycle transferring from similar datasets can still achieve a good model accuracy while the ML engineer does not have to try with many different ML configurations.

We use all large-scale object detection datasets as in table \ref{tab:detection} as well as constructing our experimental dataset as the total of three mentioned image data versions, including 57 driving videos, 2940 images. We also use the 10-center core set computation to compute the similarity of our dataset with other object detection datasets. The results are showed in table \ref{tab:detection} in the bottom row. In object detection transferring experiments, we adopt a common algorithm as Faster R-CNN \cite{fasterrcnn} with ResNet50 \cite{resnet} backbone and FPN (Feature Pyramid Network) \cite{fpn} architecture. The training epochs are kept at max 12 epochs. Other configurations will be reused from transferred ML lifecycle versions, such as data preparation, pretrained model, learning rate, and so on. Table \ref{tab:detection_results} shows the object detection results by ML lifecycle transferring. \textit{From Dataset} is the dataset that will be transferred the ML lifecycle to the \textit{Target} dataset. We also add published AutoML results as the reference since we cannot manage computing resources to reproduce the AutoML search space.

\begin{table}[ht]
  \caption{Object detection results by ML lifecycle transferring. Metric: mean Average Precision (mAP). (-): No ML lifecycle transferring, train from scratch.}
  \label{tab:detection_results}
  \begin{tabular}{|p{3.0cm}|p{1.1cm}|p{1.1cm}|p{1.5cm}|}
    \hline
    \diagbox[width=3cm]{From Dataset}{Target} & COCO & BDD & Our Dataset\\
    \hline
    From COCO &  0.374 (-) & 0.318 & 0.531 \\
    From Pascal VOC & 0.360 & 0.280 & 0.398 \\
    From BDD & 0.352 & 0.310 (-) & \textbf{0.579} \\
    From Cityscapes & \underline{0.379} & \textbf{0.335} & \underline{0.558} \\
    From KITTI & \textbf{0.389} & \underline{0.326} & 0.527 \\
    \textit{AutoML} & \textit{0.398} \cite{nas-fcos} & \textit{N/A} & \textit{N/A} \\
  \hline
\end{tabular}
\end{table}

\textbf{Results Discussion} We will discuss the \textit{relationship between dataset similarity and object detection results} by our ML lifecycle transferring. Regarding more discussion on ML lifecycle transferring for other object detection datasets, please check the appendix.

COCO dataset \cite{mscoco} is an object detection dataset for detecting general things in real-life such as chair, dog, car, person,... COCO is the most common benchmarking dataset for object detection problem. From table \ref{tab:detection} of dataset similarity, COCO is not highly similar with any other datasets with a bit closer with Cityscapes and KITTI dataset. Therefore, object detection results by ML lifecycle transferring from other datasets for COCO do not make large differences with KITTI gives the best accuracy (4\% better than no transferring) and Cityscapes achieves the second best (1.33\% better). Especially, these results do not too far from the AutoML result (only 6\% better than no transferring).

BDD dataset \cite{bdd} is a large-scale, diverse driving videos dataset with 100K images for object detection problem. It was collected from on-road driving videos so it is quite different than COCO dataset in semantic. This can be witnessed in the dataset similarity of table \ref{tab:detection} in which BDD is highly similar with other on-road datasets like KITTI and Cityscapes. These results are reflected in the object detection results (table \ref{tab:detection_results}) with Cityscapes gives the best accuracy (8\% better than no transferring) and KITTI achieves the second best (5\% better). There is no AutoML result published for the BDD dataset.

Our real-world driving image dataset was also collected from on-road driving videos in Korea. It is very similar to other driving datasets like BDD, KITTI or Cityscapes but in different locations (BDD is in the US, KITTI and Cityscapes are in Europe). The pairwise dataset distance in table \ref{tab:detection} illustrates these semantic similarity. As a result, using our ML lifecycle transferring, BDD gives the best accuracy and Cityscapes achieves the second best. Thus, our ML lifecycle transferring can help to quickly train a new dataset with a high model accuracy. An exception is in the case of COCO dataset transferring. Although COCO is not closer in the dataset similarity than KITTI, it produces a better accuracy (0.531 vs. 0.527). This exception reflects the stochasticity of ML training for large-scale, high-dimensional datasets. However, the accuracy produced by COCO does not overcome other similar datasets (e.g., BDD gives 9\% better than COCO).

Next, we will discuss on how should we \textit{choose the top-k* most similar datasets} for lifecycle transferring. From algorithm \ref{alg:transferring}, we can get the dataset similarity by each pair of system-managed datasets and new training data. If we find some highly similar datasets to our target data (by setting a threshold) as in the cases of BDD and our real-world datasets, we can choose all highly similar datasets to do ML lifecycle transferring (e.g., choose BDD, Cityscapes and KITTI in the case of our real-world dataset). If there is no clear similarity between datasets as in the case of COCO, an ML engineer can consider to train from scratch since we have no clue to determine the transferring. However, from our experimental results, one can still use our ML lifecycle transferring to get good model accuracy.

From these discussions, our ML lifecycle transferring by efficient dataset similarity computation can be used to quickly build ML lifecycle for a new object detection problem.

\subsubsection{Spatiotemporal Prediction Transferring Experiments} 

In this experimental section, we conduct experiments for spatiotemporal prediction transferring to prove our proposed solutions.

Firstly, it is required to learn embedding for each data sample of a spatiotemporal dataset using a neural network to be able to apply the core set algorithm. We leverage an \textit{autoencoder} architecture \cite{autoencoder} to train and learn the embedding for a dataset. An autoencoder is a neural network architecture that learns the representation encoding of input data by trying to reproduce the input from the embedding.

Since a spatiotemporal data can be represented as a graph \cite{dcrnn}, we implement the autoencoder model based on a graph neural network (GNN) sequence-to-sequence algorithm as in \cite{dcrnn}. The learned embedding dimensions of each spatiotemporal dataset are shown in table \ref{tab:spatiotemporal_distance} with the first dimension is the number of hidden units, and the second dimension is the number of data nodes.

We use the Gromov–Wasserstein (G-W) Distance algorithm \cite{gw-distance} to compute the distance between each pair of datasets which have different dimensions. A notable remark is that we cannot use the original number of data samples to compute distance because the G-W algorithm runs too long for large datasets. Using the k-center approximation core set (with k=100), we can run the G-W algorithm to compute the distance between spatiotemporal datasets. Table \ref{tab:spatiotemporal_distance} presents the pairwise distance between our examination datasets.

Based on the table, some groups of similar datasets can be constructed using pairwise distance values. \textit{Group 1} includes traffic speed sensor data of dataset \#1 and \#2. \textit{Group 2} consists of air pollution data of dataset \#3 and \#4. These constructed groups agree with each dataset's semantic characteristics, proving our dataset similarity computation based on the core set. Moreover, another group of similar datasets can be established is \textit{group 3} of dataset \#1 and \#3. It demonstrates that traffic speed and air pollution spatiotemporal datasets can share the similar data semantic which is intuitive.

\begin{table}[ht]
  \caption{Pairwise distance between spatiotemporal datasets}
  \label{tab:spatiotemporal_distance}
  \begin{tabular}{|p{1.cm}|p{1.5cm}|p{0.75cm}|p{0.75cm}|p{0.75cm}|p{0.75cm}|}
    \hline
    Dataset & Embedding Dimensions & \#1 & \#2 & \#3 & \#4\\
    \hline
    \#1        & 64 x 207 &  & 0.022 & \textbf{0.020} & 0.026\\
    \#2        & 64 x 325 & \textbf{0.022} & & 0.032 & 0.045\\
    \#3        & 64 x 25  & 0.020 & 0.026 &  & \textbf{0.015}\\
    \#4        & 64 x 37  & 0.032 & 0.045 & \textbf{0.015} & \\
  \hline
\end{tabular}
\end{table}

We do experiment with the ML problem as predicting traffic speed and air pollution for 12 time steps ahead from 12 time steps before. We present the experimental results of lifecycle transferring for spatiotemporal prediction by considering the following scenarios. The first scenario is to transfer the ML lifecycle of traffic speed dataset \#1 to \#2 (group 1 of similar datasets). Following the paper \cite{dcrnn}, for dataset \#1 (speed sensor data in LA city, USA), we build use data preparation as graph transformation. The model algorithm uses Diffusion Convolutional Recurrent Neural Network (DCRNN) architecture, a GNN-based spatiotemporal prediction algorithm.

Table \ref{tab:spatiotemporal_exp_1} shows various model configurations corresponding to different model versions of ML lifecycle for dataset \#1. We change the numbers of hidden units (e.g., 64 to 128) and the numbers of recurrent neural network (RNN) layers (e.g., 2 to 3) and also combinations between them. The learning rate for the training version is 1e-2. These lifecycle versions are then transferred to build ML lifecycle for traffic speed dataset \#2 (speed sensor data in the Bay area, USA) using the same data preparation, model algorithm, and learning rate of training version.

The second examination case is to transfer the ML lifecycle of the air pollution dataset \#3 to \#4 (group 2). We construct ML lifecycle versions for dataset \#3 (PM2.5 air pollution in Seoul, Korea) following the paper \cite{convlstm}, an image-based air pollution prediction solution. The model version uses the Convolutional Long Short Term Memory (ConvLSTM) model, a CNN-based air pollution forecasting algorithm. These lifecycle versions are transferred to build the ML lifecycle for dataset \#4 (PM10 air pollution in Seoul, Korea). The learning rate used for the training version is 1e-4.

The final scenario is to answer the question: is an ML lifecycle that works best for dataset \#1 also suitable for \#3 (as of the same group 3)? We experiment by transferring a DCRNN-based lifecycle version of dataset \#1 to \#3 and comparing it with ConvLSTM-based lifecycle versions. We change the learning rate for the training version to 1e-3. 

All training versions for 3 scenarios use max 100 epochs with early stop and uses mean absolute error (MAE) as the output metric.

\begin{table}[ht]
  \caption{Spatiotemporal lifecycle transferring results for traffic speed prediction (scenario 1)}
  \label{tab:spatiotemporal_exp_1}
  \begin{tabular}{|p{1.5cm}|p{4.0cm}|p{1.5cm}|}
    \hline
    Dataset (traffic speed) & Model version (DCRNN-based) & MAE\\
    \hline
    \#1 (LA) &  \textbf{Hidden units: 64 \newline RNN layers: 2} & \textbf{3.047}\\
             &  Hidden units: 128 \newline RNN layers: 2 & 3.068\\
             &  Hidden units: 64 \newline RNN layers: 3 & 3.068\\
    \hline
    \#2 (Bay) &  \textbf{Hidden units: 64 \newline RNN layers: 2} & \textbf{1.626}\\
             &  Hidden units: 128 \newline RNN layers: 2 & 1.681\\
             &  Hidden units: 64 \newline RNN layers: 3 & 1.666\\
  \hline
\end{tabular}
\end{table}

\begin{table}[ht]
  \caption{Spatiotemporal lifecycle transferring results for air pollution data (scenario 2 and 3)}
  \label{tab:spatiotemporal_exp_2}
  \begin{tabular}{|p{1.5cm}|p{4.0cm}|p{1.5cm}|}
    \hline
    Dataset (air pollution) & Model version (ConvLSTM-based) & MAE\\
    \hline
    \#3 (PM2.5) &  \textbf{LSTM 3 layers} & \textbf{10.134}\\
             &   GRU 3 layers & 10.144\\
             &   LSTM 2 layers & 10.283\\
             &   \textbf{DCRNN} & \textbf{7.270}\\
    \hline
    \#4 (PM10) &  \textbf{LSTM 3 layers} & \textbf{17.256}\\
             &   GRU 3 layers & 17.301\\
             &   LSTM 2 layers & 17.386\\
             &   \textbf{DCRNN} & \textbf{13.016}\\
  \hline
\end{tabular}
\end{table}

\textbf{Results Discussion} We will discuss the \textit{spatiotemporal prediction results} for 3 scenarios of ML lifecycle transferring. Table \ref{tab:spatiotemporal_exp_1} presents the results for the first scenario of transferring datasets in group 1 (traffic speed data). We can see that a model version configuration (e.g., 64 hidden units, 2 RNN layers) producing the best prediction result for dataset \#1 (the first row in the table) can achieve the smallest error for dataset \#2. These results claim the ML lifecycle transferring algorithm can work well for spatiotemporal traffic speed datasets.

In scenario 2, we validate how to transfer the ML lifecycle versions of two spatiotemporal air pollution datasets (group 2). Various ML lifecycle versions are illustrated in table \ref{tab:spatiotemporal_exp_2} with different model versions are different ConvLSTM architectures, such as using an LSTM \cite{lstm} or a GRU \cite{gru} model as a recurrent neural network (RNN) model, the numbers of layers are also various. From table \ref{tab:spatiotemporal_exp_2}, we can realize that a model version (e.g. LSTM 3 layers) can produce the best MAE for both two similar spatiotemporal air pollution datasets that proves our solution of ML lifecycle transferring for spatiotemporal air pollution data.

Scenario 3 examines how to transfer ML lifecycle of a traffic speed dataset to a air pollution dataset (group 3). Table \ref{tab:spatiotemporal_exp_2} proves that the DCRNN-based model version of dataset \#1 can produce the smallest MAE for both dataset \#3 and \#4. These results validate the efficiency of the ML lifecycle transferring algorithm for different spatiotemporal datasets.

\subsection{The Advantages and Limitations of our Solution}

\textbf{Advantages} Our solution reduces memory and computation to compute dataset similarity for high-dimensional, large-scale datasets by the core set-based data distance. 

Our method also decreases human effort, time and computation to build an end-to-end ML lifecycle for a new ML problem by transferring ML lifecycle versions from a group of similar datasets in the system. We prove the effectiveness and efficiency of our method in two large-scale, high-dimensional data of image and spatiotemporal. It can be helpful for other data types like text, video, or graph.

\textbf{Limitations} The first problem with our method is the dataset similarity by core set selection. The k-center greedy approximation for core set selection will select outliers in the data distribution that makes the core set computation could be unstable. Recent research suggests a Probability Coverage solution \cite{ProbCover} for the core set selection. An another concern with the core set algorithm is how to choose the right number of k centers in the greedy k-center algorithm. One solution could be choosing k as the number of classes in the dataset.

The next limitation of our solution is the method of transferring lifecycle versions to a new training dataset. The question is whether it is too simple to just transfer lifecycle versions’ configurations? How can we add more optimization in model or training during the lifecycle transferring? And the last one is how can we use this solution to improve AutoML methods which also try to automatically learn an ML pipeline for a new training data but in a huge search cost?

One last restriction of our solution is it has not been used by many ML teams to get quantitative feedback.

\section{ML Lifecycle Rebuilding}
In this section, we present how VeML supports various ML lifecycle rebuilding methods with a new testing data version. As mentioned in the introduction, when the testing data is dissimilarity in data distribution with the training data, model performance will degrade and we will need to rebuild the ML lifecycle. Thus, in the first part, we will introduce how to detect data distribution mismatch between pairs of data versions in our system.

\subsection{Data Distributions Mismatch Unsupervised Detection}

We propose an algorithm to automatically detect data distribution dissimilarity between a testing data version and the training data version without getting labeled data. Our method has two stages as follows. 

First, we use the core set and covering balls to represent the data distribution of each data version, as mentioned in the core set computation section (figure \ref{fig:core_set_algorithm}). Second, we compute the distance between each data point in the core set of a testing data version to the nearest center point of the training data version's covering balls. We employ this computation to examine whether the core set of a testing data version is inside the covering balls of the training data version, which can decide their dissimilarity in data distributions. 

\begin{figure}[ht]
  \centering
  \includegraphics[width=0.8\linewidth]{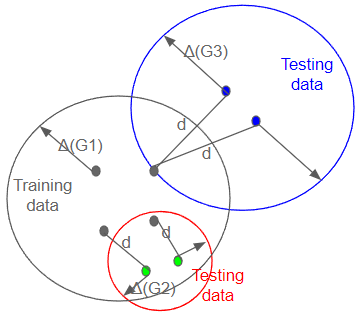}
  \caption{Data distribution of training and testing data versions.}
  \label{fig:data_distribution}
\end{figure}

Suppose the average of these distances is less than the covering distance of the training data version. In that case, the core set of a testing data version is covered by covering balls of the training data version. Thus, we can conclude that the testing data version's data distribution follows the data distribution of the training data version. In contrast, the testing data version's core set is outside the training data version's covering balls, indicating the testing data is drawn from a different data distribution than the training data. Figure \ref{fig:data_distribution} illustrates the data distribution of training and testing data versions and how to detect their dissimilarity using our method. $\Delta(G)$ is the covering radius of a data version, and $d$ is the distance between a data point in a data version to the nearest core set center of another data version.

Consequently, our solution can work without labeled data (since the core set algorithm does not depend on data labels) and can automatically detect data distribution mismatch between testing and training data versions. The next section will convince our algorithm by experimenting on data distribution mismatch detection with a real-world driving dataset.

\subsection{Distribution Dissimilarity Detection Experiments}

This section show our experiments to detect data distribution differences between a testing and a training data version. Three real-world image data versions described in table \ref{tab:exp_data_versions} will be used to validate our solution. We follow the greedy algorithm to compute the approximated core set $G$ and covering radius $\Delta(G)$ for each data version.

According to the algorithm for data distribution mismatch detection, we calculate distances between each data point in the core set of a testing data version and the nearest center of the training data version's covering balls. Next, we compare the average of these distances to the covering radius $\Delta(G)$ of the training data version. Table \ref{tab:data_drift} shows the results of comparing distributions between different driving data versions using our algorithm. Driving data versions in the first column act as the training data, and data versions in the first row play as the new testing data. $G$ and $\Delta(G)$ are computed using a 10-center greedy algorithm.

In table \ref{tab:data_drift}, (+) indicates a testing data version's data distribution is not covered by the training data, which means they are different in data distribution. (-) implies the training data covers the distribution of the testing data, or testing data are drawn from the same data distribution as the training data. From the table, only the testing data D1018 is covered by the distribution of the training data D0114 (since $d$=5.81 < $\Delta(G)$=6.85), while other pairs of data versions are different in data distribution.

\begin{table}[ht]
  \caption{Comparing data distributions using the core set. Data versions in the first column act as training data. Data versions in the first row play as testing data}
  \label{tab:data_drift}
  \begin{tabular}{|p{1.cm}|p{1.cm}|p{1.5cm}|p{1.5cm}|p{1.5cm}|}
    \hline
    Data version & $\Delta(G)$ (k=10) & D0821 & D1018 & D0114\\
    \hline
    D0821 & 5.69 & & 6.88 > $\Delta(G)$ (+) & 7.42 > $\Delta(G)$ (+)\\
    D1018 & 5.55 & 7.39 > $\Delta(G)$ (+) & & 6.71 > $\Delta(G)$ (+)\\
    D0114 & 6.85 & 7.20 > $\Delta(G)$ (+) & \textbf{5.81} < $\Delta(G)$ (-) & \\
  \hline
\end{tabular}
\end{table}

\textbf{Experimental results} We sequentially experiment with each data version as the training and testing data to prove our unsupervised data distribution detection algorithm. We adopt the same ML pipeline for all of our trails, validation data is splitted from the training data to make them have the same distribution.

The experimental results are illuminated in table \ref{tab:data_drift_exp}, on which we can observe that if training on data version D0821 and testing on D0114 or D1018, the model accuracy degrades more than 20\% from validation to testing accuracy. Whereas, when training on data version D0114 and testing on D1018, the model accuracy drops at only 10\%. These results confirm our previous analysis that data version D1018 can be considered to have no difference in the data distribution compared to D0114 and, thus, little model accuracy drops.

On the other hand, data versions D1018 and D0114 have a significant dissimilarity in the data distribution compared to data version D0821; hence, their model accuracy dropped quite large (2x larger) which requires a model retraining.

\begin{table}[ht]
  \caption{Object detection results with pairs of training and testing data versions. Metric: mean Average Precision (mAP).}
  \label{tab:data_drift_exp}
  \begin{tabular}{|p{1.cm}|p{1.cm}|p{1.5cm}|p{1.5cm}|p{1.5cm}|}
    \hline
    Training data & Testing data & Validation Accuracy & Testing Accuracy & Accuracy Drop\\
    \hline
     D0821 & D1018 & 0.723 & 0.546 & 0.177 (24\%)\\
     D0821 & D0114 & 0.689 & 0.545 & 0.144 (21\%)\\
     D0114 & D1018 & 0.695 & 0.619 & \textbf{0.076 (10\%)}\\
  \hline
\end{tabular}
\end{table}

\subsection{ML Lifecycle Rebuilding Methods}

From above experimental results with pairs of training and testing data versions, we need to retrain the model when a testing data is significantly different than the training data, which causes the ML lifecycle rebuilding.

To support ML lifecycle rebuilding with a new testing data version, VeML implements various model retraining algorithms to favor this requirement. This section presents several retraining algorithms, their advantages (pros), disadvantages (cons), and the ability to be automated implementation.
\begin{itemize}
    \item \textbf{Full Training} works by retraining a new ML model with all data every time a new testing data coming. Pros: highest testing accuracy, easy to automate. Cons: very high training time and computing consumption, needs labeled data for all data samples.
    
    \item \textbf{Transfer Learning} is to retrain a new ML model with only new data from a previous pretrained model. Pros: faster in training, high testing accuracy, easy to automate. Cons: catastrophic forgetting problem, still needs labeled data for every new data samples.
    
    \item \textbf{Domain Adaptation} is retraining with both labeled (from a source domain) and unlabeled (from a target domain) data. Pros: using only unlabeled data of the target domain. Cons: problem-dependent domain adaptation algorithms, hard to automate.
    
    \item \textbf{Active Learning} selects the most informative data points for labelling from the new unseen data to do model retraining. Pros: reducing the effort to annotate label data. Cons: problem and data dependent active learning algorithms, hard to automate.
    
\end{itemize}

Based on above mentioned incremental training methods, we implement ML lifecycle rebuilding on VeML for a new testing data version. Firstly, we introduce the ML lifecycle versions settings in our system. Assume that the ML problem has run through many cycles with some training data versions, denoting as $d_{1}$, $d_{2}$,... $d_{k}$, and a new testing data version denoted $d*$. ML lifecycle versions also consist of a sequence of model versions, $m_{1}$, $m_{2}$,... $m_{n}$, and training versions, $t_{1}$, $t_{2}$,... $t_{p}$.

VeML implements three common methods to rebuild a ML lifecycle for the new testing data version $d*$ as follows:
\begin{itemize}
    \item \textbf{Full Training Method.} An ML engineer labels all data points in $d*$ and creates a full training data version $d’$ = merge ($d_1$, $d_2$,..., $d_k$, $d*$) in our system. Then, he or she creates a new model version $m*$, a new training version $t*$ from $m_n$ and $t_p$ versions, respectively (since full training method uses the same model architecture and training hyper-parameters for retraining). Next, VeML can do model retraining to rebuild a new ML lifecycle from these lifecycle versions.
    
    \item \textbf{Transfer Learning Method.} An ML practitioner selects a model version $m*$ from $m_n$ (same model architecture as the previous model version), and a training version $t*$ from $t_p$ (reusing trained model of the previous training version). VeML will train $m*$ on the training data version $d*$ (labeling needed) with training version $t*$ and rebuild a new ML lifecycle.
    
    \item \textbf{Active Learning Method.} An ML engineer chooses an active learning algorithm to select the most informative data points from the new data version $d*$ to label. Then, he or she creates a new training data version $d’$ = merge ($d_1$, $d_2$, … $d*$(active$\_$learning)). VeML will train a model version $m*$ that is the same architecture as $m_n$ version on $d’$ and rebuild the ML lifecycle.
    
\end{itemize}

In the next section, we will present a real-world scenario for a self-driving project that illustrates how VeML supports ML lifecycle rebuilding with a new data version.

\subsection{Experimental Results}

This section shows experiments for incremental training on driving image dataset \ref{tab:exp_data_versions}. We consider a real-world scenario for the self-driving project as follows. An ML engineer for the self-driving project collects and labels image dataset in a day of driving to build an object detection model. Data version D0821 which includes driving videos in the day 08/21 will be the training data version. After building a production object detection model, the self-driving ML engineer collects new driving videos in another day. An important question is whether the production model still works well for new data or we need to retrain the model? Data version D1018 which includes many driving videos in the day 10/18 will be the new data version. From previous experiments, VeML can detect that the new data version D1018 has different data distribution than D0821 version, which suggests the ML engineer a model retraining task.

Table \ref{tab:incremental_learning} illustrates three model retraining methods that are supported in our system: full training, transfer learning, and active learning. In the active learning method, we use the algorithm in the paper \cite{coreset} that chooses the numbers of data samples to label following the core set computation. An ML engineer can use an another active learning algorithm to select a small number of labeled data points. In our case, we do experiments with different ratios of labeled data, such as 10\%, 30\%, and 50\% of the whole data points.

\begin{table}[ht]
  \caption{Experimental Results for Model Retraining on Our Image Dataset. Metric: mean Average Precision (mAP).}
  \label{tab:incremental_learning}
  \begin{tabular}{|p{2.cm}|p{2.5cm}|p{1.cm}|p{1.cm}|}
    \hline
    Method & \#Labeled Data Needed & Testing Accuracy & Training Time (minutes)\\
    \hline
      No retraining & - & 54.55 & - \\
      \hline
      Full training & 673 & \textbf{59.24} & 65\\
      Transfer learning & 673 & 58.40 & \textbf{20}\\
      Active Learning & 67 (10\% data points) & 56.83 & 48\\
                      & 201 (30\% data points) & 58.12 & 51\\
                      & 336 (50\% data points) & \underline{58.57} & 56\\
  \hline
\end{tabular}
\end{table}

\textbf{Results Discussion} Firstly, we discuss about the \textit{effectiveness of retraining methods}. From the experimental results, when we do not execute retraining, the testing accuracy is quite low (54.55\% mAP). When we execute model retraining, the full training method produces the highest testing accuracy upgrade compared to no retraining (8.6\% better). Active learning method with only 50$\%$ labeled data can achieve the second-highest testing accuracy (7.4\% better). The transfer learning has the smallest training time but it still needs a lot of labeled data compared to the active learning method (673 vs. 336 label data points).

Secondly, we argue on \textit{which model retraining approach is better} and the ability of VeML for automation. Each model retraining method would have its pros and cons, and is helpful in different context. For example, transfer learning method is common used in model retraining because of its fast training but it has the problem of catastrophic forgetting, the tendency of a neural network to substantially forget previously learned information upon learning new information in incremental learning. In the other hand, full training method is high computation but it produces the highest model accuracy and it should be used periodically to avoid the problem of old information forgetting. Active learning is a promising method in reducing annotation effort but its algorithm depends on the specific data and ML tasks. Therefore, VeML allows the ML engineer to choose from one of supported methods and then it can \textit{automatically rebuild ML lifecycle} after the ML engineer supplies the labeled data (for the new data version). In the future, we can find how to automatically evaluate each method for a specific data and ML task and suggest for the ML engineer.

\section{VeML in a self-driving project and new ML lifecycle challenges support}

This section presents how VeML is using in an on-going self-driving project with new data versions coming continuously. Moreover, we introduce a new ML lifecycle challenge named model error track-and-trace that VeML can support in the future.

\subsection{VeML in a Self-driving Project}

We are working in a real-world self-driving project in Korea. The driving videos are collected from some mounted cameras (1 to 3 cameras) on a driving car. From these driving videos, we will extract to many frames (images) and annotate them to 30 on-road objects, such as Vehicle\_Car, Vehicle\_Bus, Pedestrian, Road-Mark, Traffic-Light,... The autonomous car project needs to deal with various object detection models, such as vehicles detection, pedestrians detection, traffic lights detection, and so on. It also requires to deploy the inference model in different environments such as servers, edge devices. Thus, VeML is an appropriate system to support us work on this real-world ML application.

VeML supports this project by managing driving images and annotations as training data versions. Using ML lifecycle transferring on VeML, we can quickly build new ML lifecycle for this project based on common object detection datasets. Moreover, when the project is running, new driving videos are continuously coming in many different driving situations: locations, weather, timeof day,... that creates many new data versions. It makes VeML a suitable tool for supporting rebuilding ML lifecycle for new data versions.

\begin{figure}[ht]
  \centering
  \includegraphics[width=\linewidth]{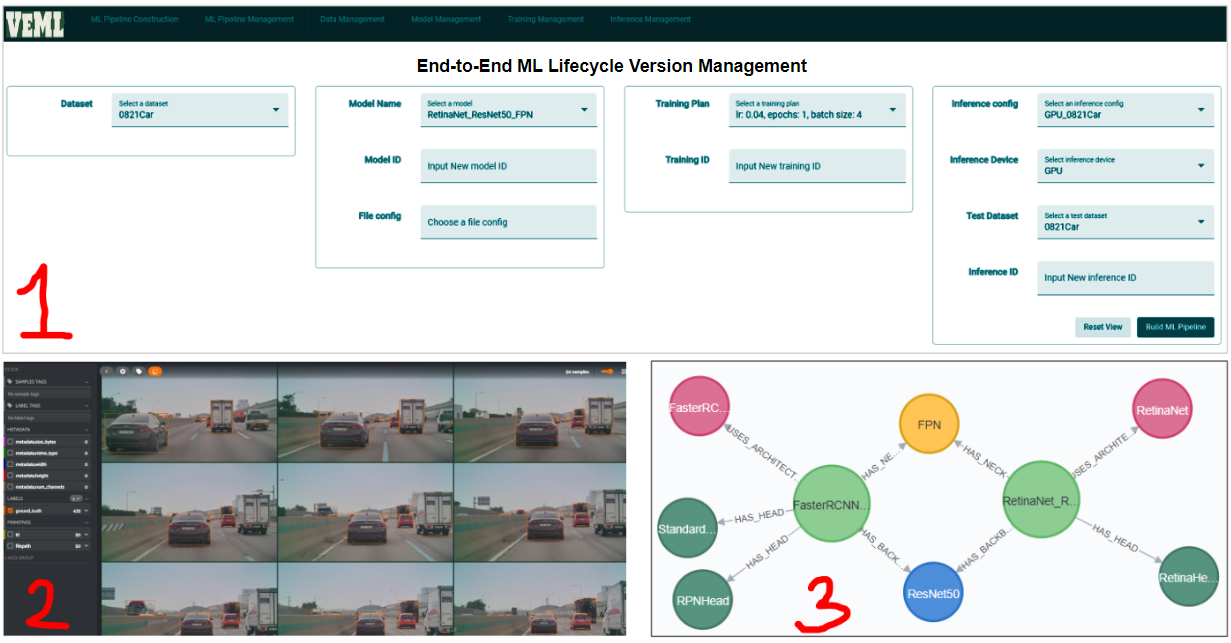}
  \caption{VeML in our working self-diving project. Box 1: Interactive User Interface for end-to-end ML lifecycle version management. Box 2: Driving image dataset visualization and analysis. Box 3: Graph-based model architectures management.}
  \label{fig:veml_mlops}
\end{figure}

Figure \ref{fig:veml_mlops} shows how VeML is using in our working real-world self-driving project. Box 1 is the main interactive User Interface (UI) for end-to-end ML lifecycle version management. It allows us to visualize the end-to-end ML lifecycle versions, from datasets, model architectures, training plans, and inference configurations. Box 2 shows the UI for dataset visualization and analysis. It can show all data samples and annotations in a unified UI for us to validate and analyze the training data. Box 3 is a UI for the graph-based management of model versions. It presents relationships between many model architectures such as model types, learning algorithms, model backbones, and so on. We can do various model versions comparison and analysis using the graph-based management.

\subsection{Model Error Track-and-Trace}

Model error Track-and-Trace is a challenge but important problem for an ML lifecycle. Track-and-Trace means to track from training data through the model training and model deployment, then when a model error occurs, we can trace back to its root causes (by the training process, by model architecture, or by training data). The ML lifecycle of an ML application can be very long and complicated so the track-and-trace function becomes more and more critical. VeML can support this capability in the end-to-end ML lifecycle by learning a Track-and-Trace model at the same time with the ML model. When the ML model generates the wrong prediction, we can use the Track-and-Trace model to find out what the problems cause it: data, features, model, or bias,...

\begin{figure}[ht]
  \centering
  \includegraphics[width=\linewidth]{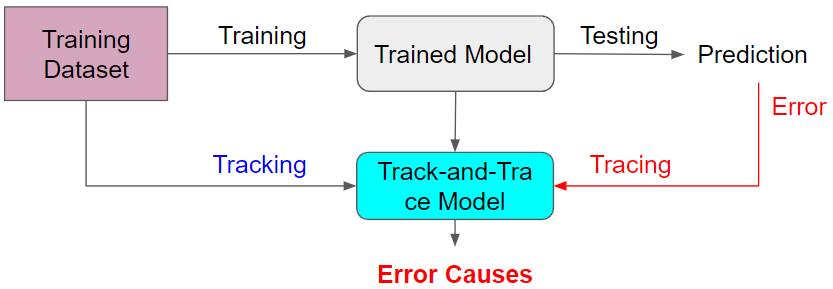}
  \caption{A framework for model error track-and-trace.}
  \label{fig:track_and_trace}
\end{figure}

Figure \ref{fig:track_and_trace} illustrates a framework for model error track-and-trace that can be supported in our VeML system. The track-and-trace model learns from training data knowledge like data statistics, data features, and learns from the trained model information like model architecture, model features, and uses these information to find the error root causes.

\section{Conclusion}

This research presents VeML, a version management system dedicated to the end-to-end ML lifecycle. We constructed our system from scratch over three main blocks, in-memory storage for large-scale data storage and logging, a graph database for graph-based version management, and an open-source training platform for ML training.

We propose two algorithms based on the core set for large-scale, high-dimensional data. The first one is a dataset similarity algorithm that can be used to transfer ML lifecycle versions of the similar dataset managed in our system to a new training data for effectively and efficiently building a new ML lifecycle. Our extensive experimental results on two large-scale, high-dimensional datasets, driving image dataset of a self-driving project and spatiotemporal sensor data, proves our proposed solution.

The second algorithm is an unsupervised data distribution mismatch detection between the testing and training data. When detecting data distribution dissimilarity, our system allows data scientists to select from various model retraining methods then it will automatically rebuild a new ML lifecycle after that.

In addition, we show how VeML is using in a real-world self-driving project to build an end-to-end ML lifecycle and works with new data versions continuously coming. VeML can also support new challenges in ML lifecycle such as model error track-and-trace.

In future work, this research can unlock many open issues and challenges in ML lifecycle problem. Efficient ML lifecycle building is a crucial problem for any ML lifecycle management system. We hope that continual research on this will open many new opportunities. Active learning is a recent field of research that reduces the number of labels data for model retraining which will be an important research for the ML lifecycle. Moreover, Human-in-the-Loop for ML lifecycle that integrates human knowledge into ML lifecycle is also a promising future research direction.

\appendices
\section{\break Object Detection Results by ML Lifecycle Transferring}

This appendix shows object detection results by ML lifecycle transferring on three target datasets: Cityscapes \cite{cityscapes}, KITTI \cite{kitti}, and Pascal VOC \cite{PascalVoc}. The object detection model is Faster R-CNN with ResNet50 backbone and FPN (Feature Pyramid Network) architecture and training for max 12 epochs.

\begin{table}[ht]
  \caption{Object detection results by ML lifecycle transferring. Metric: mean Average Precision (mAP). (*): No ML lifecycle transferring, train from scratch.}
  \label{tab:detection_results_1}
  \begin{tabular}{|p{3.0cm}|p{1.1cm}|p{1.1cm}|p{1.5cm}|}
    \hline
    \diagbox[width=3cm]{From Dataset}{Target} & Cityscapes & KITTI & Pascal VOC\\
    \hline
    From COCO &   0.331 & 0.901 & \underline{0.816} \\
    From Pascal VOC &  0.255 & 0.858 & 0.804 (*) \\
    From BDD &  \underline{0.338} & 0.893 & 0.774 \\
    From Cityscapes &  \textbf{0.406} (*) & \underline{0.903} & 0.797 \\
    From KITTI &  0.337 & \textbf{0.904} (*) & \textbf{0.825} \\
  \hline
\end{tabular}
\end{table}

\textbf{Results Discussion} Cityscapes dataset \cite{cityscapes} is an image dataset focusing on semantic understanding of urban street scenes but it still has object detection annotations for 5000 images. From table \ref{tab:detection} of dataset similarity, Cityscapes  is much closer than BDD and KITTI compared to other datasets. Then, as in table \ref{tab:detection_results_1}, object detection results by ML lifecycle transferring from BDD and KITTI have higher accuracy than from COCO and Pascal VOC but not much. Moreover, training from scratch on the Cityscapes dataset gives the highest accuracy. The reason could be that Cityscapes has a manual frames selection to be large number of dynamic objects, varying scene layout, and varying background that makes it specific.

KITTI dataset provides object detection and object orientation estimation benchmark consists of 7481 training images and 7518 test images \cite{kitti}. It was collected from on-road driving videos so it is more similar with BDD and Cityscapes than COCO and Pascal VOC as showed in table \ref{tab:detection}. Regarding the object detection results by ML lifecycle transferring from table \ref{tab:detection_results}), Cityscapes gives the second best accuracy while training from scratch with KITTI achieves the best result. Specially, transferring from COCO produces a slightly better accuracy than from BDD dataset (0.901 vs. 0.893, a 1\% better). This case suggests COCO is a good base dataset to do lifecycle transferring for other datasets.

The Pascal VOC or VOC dataset provides standardised image data sets for object class recognition \cite{PascalVoc}. It has 20 general classes like car, bus, person, cat, chair,... with the train/val data has 11,530 images. From dataset similarity \ref{tab:detection}, Pascal VOC is not significant closer than any other datasets. The object detection results show that a general object dataset like COCO can give a high model accuracy (second best) by lifecycle transferring. The best accuracy achieved by transferring from KITTI suggests that when we do not find highly similar datasets, we cannot determine the ML lifecycle transferring and training from scratch could be a suitable option.

\begin{IEEEbiography}[{\includegraphics[width=1in,height=1.25in,clip,keepaspectratio]{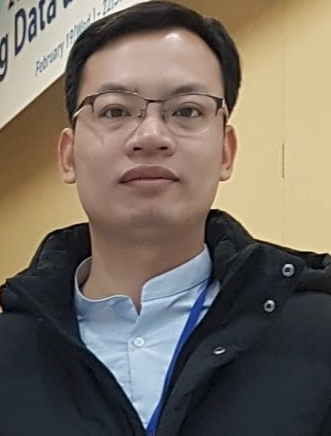}}]{Van-Duc Le} is a Ph.D. candidate at School of Electrical and Computer Engineering, Seoul National University, Seoul, South Korea. His research interests include Spatiotemporal Deep Learning, Ambient AI, and Machine Learning Lifecycle Management. He received his Master Degree from Seoul National University, Korea in 2019.
\end{IEEEbiography}

\begin{IEEEbiography}[{\includegraphics[width=1in,height=1.25in,clip,keepaspectratio]{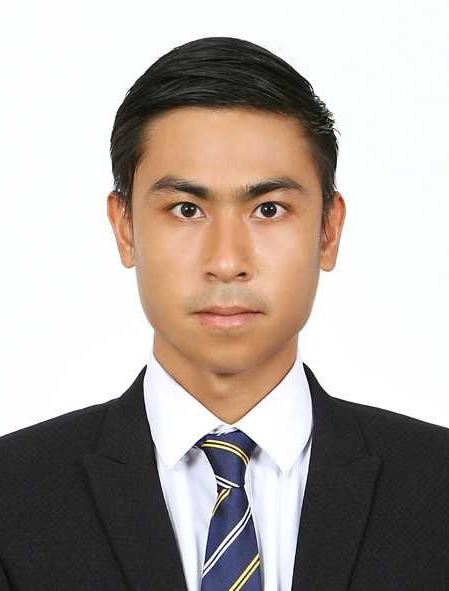}}]{Tien-Cuong Bui} is a Ph.D. candidate at School of Electrical and Computer Engineering, Seoul National University, Seoul, South Korea. His research interests include Data Mining, Natural Language Processing, Graph Mining, and Intelligent Infrastructure. He received his Master Degree from Seoul National University, Korea in 2019.
\end{IEEEbiography}

\begin{IEEEbiography}[{\includegraphics[width=1in,height=1.25in,clip,keepaspectratio]{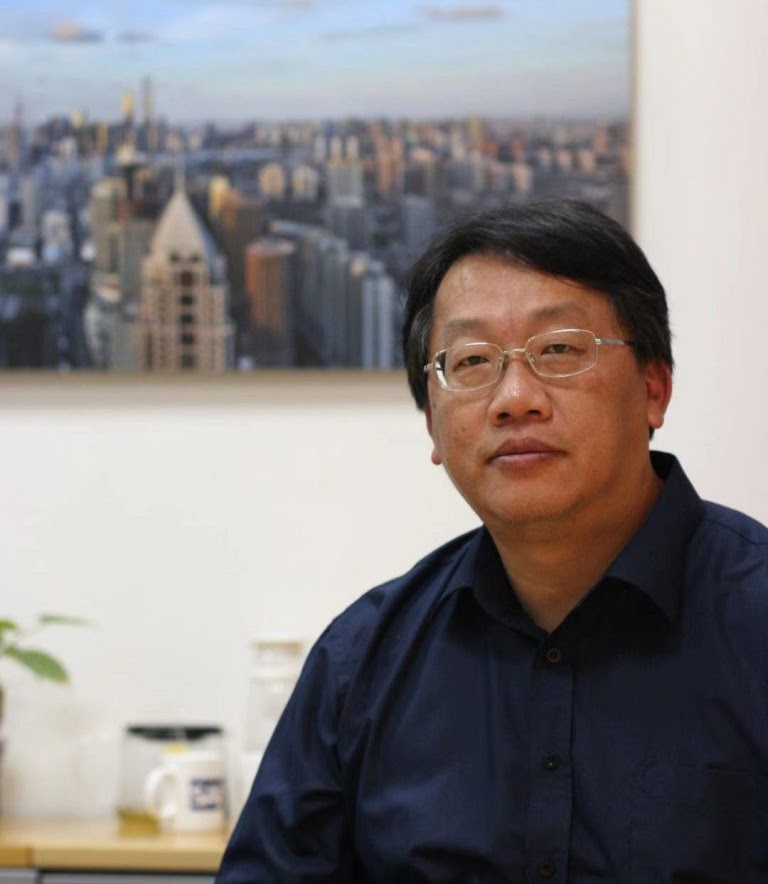}}]{Dr. Wen-Syan Li} joined the Graduate School of Data Science, Seoul National University (SNU) as a Full Professor in March 2020 and became a Foreign Fellow of the Brain Pool Program under the National Research Foundation of Korea in June 2020. Before joining SNU, he was Senior Vice President of SAP SE and Head of SAP Customer Innovation and Strategic Projects – Asia Pacific, Japan, and Greater China. His team worked on the new applications in the area of digital supply chain and strategic engagements with key accounts such as Huawei, NTT, Intel, and Lenovo in the area of IoT and SAP Hana. His team was also responsible for building Predictive Analytics capabilities in SAP’s in-memory database HANA. 

He received a Ph.D. degree in Computer Science from Northwestern University (USA). He also has an MBA degree in Finance. Before joining SAP, he was with IBM Almaden Research Center located, NEC Research, and NEC Venture Capital in the USA. He has co-edited 3 books published by Springer, co-authored more than 100 journal articles and conference papers in various areas, and co-invented 82 granted US patents. 

His research interests include ML/DL, human-in-the-loop AI, machine learning life cycle management and automation, big data and knowledge management, and applying machine learning on solving real-world problems.
\end{IEEEbiography}

\EOD

\end{document}